\documentclass[report]{IEEEtran}
\usepackage{graphicx}
\usepackage{grffile}
\usepackage[]{caption}
\usepackage{amsmath}
\usepackage{amssymb}
\usepackage{algorithm}
\usepackage{subcaption}
\usepackage{color}
\usepackage[dvipsnames]{xcolor}
\usepackage{cancel}
\usepackage{float}
\usepackage{setspace}
\usepackage[normalem]{ulem}

\makeatletter
\let\@currsize\normalsize
\makeatother

\newdimen\figrasterwd
\figrasterwd\textwidth

\newcommand{\etal}{\textit{et al.}}

\newcommand{\citethis}[1][]{\textcolor{red}{cite}}
\newcommand\ignore[1]{}

\begin{document}
\title{Hefty: A Modular Reconfigurable Robot for Advancing Robot Manipulation in Agriculture}

\author{
Dominic~Guri,
Moonyoung~Lee,
Oliver~Kroemer,
George~Kantor
}

\maketitle

\begin{abstract}
This paper presents a modular, reconfigurable robot platform for robot manipulation in agriculture. While robot manipulation promises great advancements in automating challenging, complex tasks that are currently best left to humans, it is also an expensive capital investment for researchers and users because it demands significantly varying robot configurations depending on the task. Modular robots provide a way to obtain multiple configurations and reduce costs by enabling incremental acquisition of only the necessary modules. The robot we present, Hefty, is designed to be modular and reconfigurable. It is designed for both researchers and end-users as a means to improve technology transfer from research to real-world application. This paper provides a detailed design and integration process, outlining the critical design decisions that enable modularity in the mobility of the robot as well as its sensor payload, power systems, computing, and fixture mounting. We demonstrate the utility of the robot by presenting five configurations used in multiple real-world agricultural robotics applications.
\end{abstract}

\section{Introduction}\label(sec-intro)
With global warming and an increasing population, the world needs to improve food production, and agricultural robotics could be an important part of the solution. In addition to mitigating labor shortages and relieving workers from strenuous and dangerous tasks, robotics may also help the adoption of ``Agriculture 4.0''\cite{oliveiraAdvancesAgricultureRobotics2021}, an automation-driven, environmentally-friendly farming approach expected to minimize waste and increase yields. Specifically, using robots in farming will improve the data collection and processing necessary for better decision-making. Additionally, with improvements in perception and manipulation, robotics also provides the machinery for executing complex tasks like targeted pesticide administration, tree pruning, and fruit and vegetable harvesting -- this will be particularly essential for labor-intensive high-value crops like grapes. 

Broadly adopting robotics-based infrastructure is a significant capital investment that is yet to be affordable. However, roboticists and researchers developing this infrastructure could learn from existing tractor-based mechanization that significantly reduces costs by using a single mobile platform (the tractor) that uses a modular interface to attach various tools. Thus, by using a tractor, farmers can execute numerous brute-force tasks like tilling, disking, and hauling, provided they use the right tool. Extending such modularity to robotics infrastructure makes it affordable to both end-users and researchers; thus, it positively impacting its development and adoption.

An effective robot design combines the best mechanical form, computational and sensor hardware, and algorithmic capabilities to match the task requirements. In agriculture, all three factors are essential for the robot to perceive the environment, navigate, and manipulate sensitive crops and plants -- all facilitated by capable algorithms. The mechanical form of the robot includes the mobile base, the manipulators that interact with the environment, and the whole physical structure that makes the robot's geometry. The shape and mounting of the robot's computers and sensors also affect the mechanical form of the robot. The modularity of the robot system refers to the design of independent mechanical components so they are easily combined when building the robot, and reconfigurability allows the components to be rearranged to adapt the robot to the task requirements. Combined modularity and reconfigurability can improve repairability and reduce acquisition costs.

\begin{figure*}[ht]
  \centering
  \includegraphics[width=0.9\textwidth]{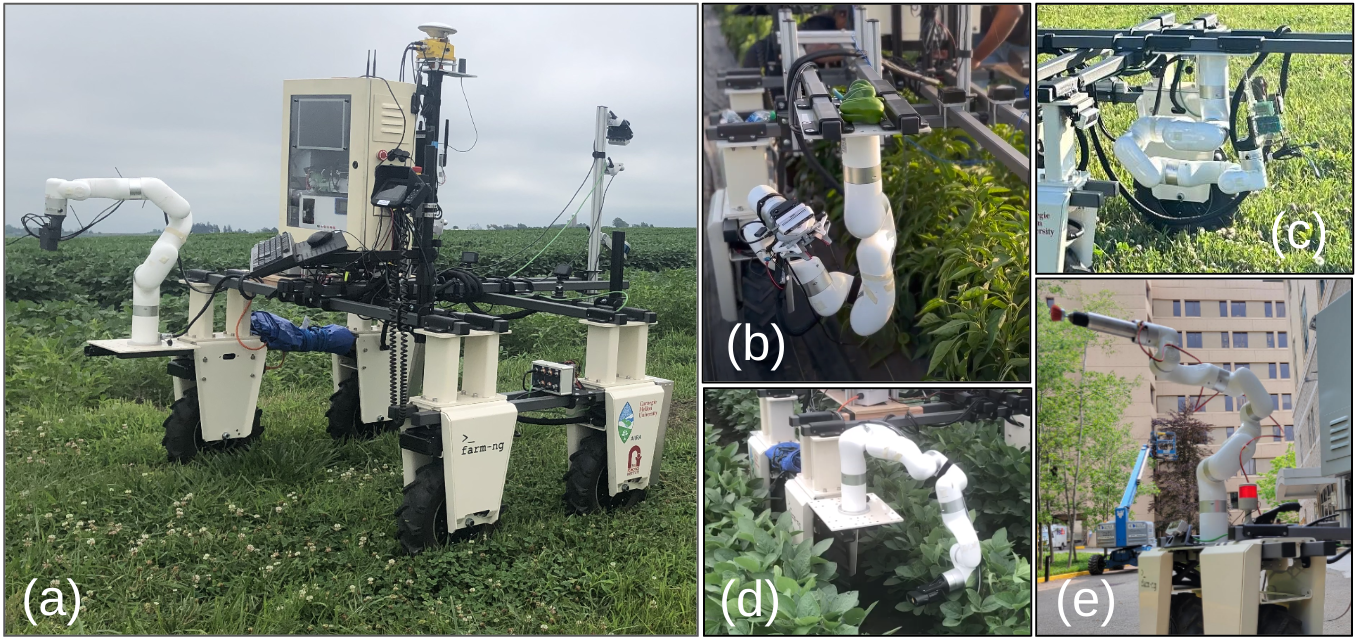}
  \caption{A complete view of Hefty in (a) also shows the keyboard, mouse, and monitor for interfacing with the system computer. The black mast carries the GPS and IMU sensors, and the aluminum has the perception sensors: a 3.3MP RGB camera, a PoE stereo camera, a Realsense camera, and a Lidar. Subfigures (b)--(e) showcase the various manipulator configurations employed in (b) pepper harvesting, incorporating a modified structure to elevate the mounting plate; (c) cornstalk sensor insertion; (d) soybean insect scouting; and (e) lantern-fly egg mass decimation.}\label{fig:tasks}
\end{figure*}

\section{Related Works}

The need for utility robot platforms for agricultural is motivated by the growing number of published works that demonstrate that robots have become effective in almost all farming operations, and continue to improve.
Even tasks like seeding and transplanting, which are extensively mechanized through tractor-based equipment, still present challenges that robotics can contribute to fixing.
For example, Srinivasan \etal \cite{srinivasanDesignAutonomousSeed2016} developed a compact (dimensions 0.6m$\times$0.5m$\times$0.3m) rugged  robot, with a tank treads drive system, for seeding even in undulating terrains -- thus, potentially extending to seeding for reforestation.
Haibo \etal \cite{haiboStudyExperimentWheat2015} introduced an all-wheel-steerable four-wheel drive robot for high precision wheat seeding. 

Indiscriminate herbicide applications is the predominant weeding method in commercial farming. And the consequences for the environment are the key motivations for robotics-based weeding. 
Robotics-based weeding typically uses targeted herbicide application or mechanical weeding. Targeted herbicide application minimizes waste and the negative environmental impact. 
Examples of targeted herbicide application robots include Ladybird\cite{benderHighresolutionMultimodalData2020} from University of Sydney or  
Ecorobotix's \textit{ARA} \cite{EcorobotixSmartSpraying}.
Ovchinnikov \etal's weed cutting robot \cite{ovchinnikovKinematicStudyWeeding2019}, and BoniRob\cite{ruckelshausenBoniRobAutonomousFielda}, a robot that kills weeds by stomping them into the ground, are good examples of mechanical weeding robots. 
The \textit{laserweeder}, from Carbon Robotics\cite{CarbonRobotics}, uses lasers to target weeds and excess plants, making it a weeding and crop thinning robot.
Modularity and reconfigurability are not widely exploited in designing robots for weeding. 
Only the Ladybird and BoniRob are reported as multi-purpose robots.
The commercially available robots are particularly task-specific, hence they contribute to making the adoption of robotics technology more cost prohibitive.

The bulk of harvesting robots prioritizes high-value fruits and vegetable because they require selective harvesting\cite{vishnurajendranSelectiveHarvestingRobots}. 
Selective harvesting requires multiple harvests that are determined by factors like the produce's ripeness, time of day, and the plant's fruiting schedule, and it generally requires meticulous handling of both the plants and the fruits.
Some of the notable robot harvesting works include manipulator and end-effector designs for apple harvesting\cite{silwalDesignIntegrationField2017,davidsonProofofconceptRoboticApple2016}, eggplants \cite{hayashiRoboticHarvestingSystem2002,sepulvedaRoboticAubergineHarvesting2020}, and strawberries\cite{yuRealTimeVisualLocalization2020} among many crops \cite{bacHarvestingRobotsHighvalue2014,vishnurajendranSelectiveHarvestingRobots}.
The diversity in plant architecture and handling requirements suggests that a single manipulator design is not suitable for all agricultural manipulation tasks, and this is particularly true for harvesting. 
Modular manipulators would benefit the users by making it possible to construct the manipulator fit for the task. 
Pfaff \etal\cite{schutzModularRobotSystema} explored modular robot arm design for harvesting sweet-peppers, apples, and grapes in different greenhouse settings.
Works from Levin \etal \cite{levinDesignTaskBasedModular2016,levinConceptualFrameworkOptimization2019} explore the optimal manipulator design problem that arises when designing a reconfigurable manipulator. 

Tasks like pruning, phenotyping, and inspection require sophisticated perception and manipulation. Plant phenotyping characterizes plant varieties through a wide range of qualitative and quantitative measurements of biochemical, morphological, and structural features, amongst many other plant traits.
Robot-based phenotyping unlocks high throughput data collection and processing that enables better analysis, including via machine learning\cite{atefiRoboticTechnologiesHighThroughput2021}. A number of works have been published collecting physical measurements like plant height, stalk width, and leaf size and shape\cite{zhangFieldPhenotypingRobot2016,xiangFieldBasedRobotic2023,baoFieldbasedArchitecturalTraits2019,alenyaRoboticLeafProbing2012}. Other works have focused on biochemical characteristics like chlorophyll content\cite{atefiVivoHumanlikeRobotic2019,alenyaRoboticLeafProbing2012}. The Robotanist\cite{mueller-simRobotanistGroundbasedAgricultural2017,abelInfieldroboticLeafGrasping2018} is an example a phenotyping robot designed for collecting physical measurements as well as chlorophyll content. 

Much of the progress discussed above has not been widely adopted partly due to the fragmentation of the robot platforms on which the technologies are presented.
However, recently more works have been dedicated to developing a holistic robotic system for general farming operations.
The \textit{Agri.q} robot \cite{colucciKinematicModelingMotion2022} is an example of a utility robot for
general agricultural applications. It is a dual-bogie (eight-wheel) drive mobile system with a reconfigurable
mounting for its 7DOF robot arm and body articulation for pitching. However, it primarily designed for vineyard plant health monitoring.
Nonetheless, it is an excellent example of an integrated utility robot tailored for autonomous navigation and diverse manipulation within
agricultural contexts; it is also used as a UAV docking station.
Another example of utility research robot is Bumblebee designed by Silwal \etal \cite{silwalBumblebeePathFully2022}. 
This design extends the \textit{Warthog} mobile robot, from Clearpath Robotics Inc., into a mobile manipulation robot capable of autonomous navigation, inspection, and pruning, and it has been used in vineyard and orchard applications.

High-value crops often require significant different growing conditions, from plant spacing, to trellising styles. As such, robots should be designed to fit the farming conditions. The Thorvald-II robot systems, developed by Lars \etal~\cite{grimstadThorvaldIIAgricultural2017}, is a modular robot platform that allows the farmer to reconfigure the robot to meet the farm requirements.
The Thorvald-II system isolates the complex robot structures like steering and suspension into modules located next to each other, typically on one side of the robot. The frame that connects two sides of the robot is the only component that changes based on task requirements like height clearance and bed width.
Farm-ng also designed their Amiga robot\cite{Farmng} to be similarly modular and reconfigurable. Thorvald-II and Farm-ng have successfully commercialized their products, which could be a promising sign of the advent of modular robot platforms that will help transfer robotics technology research to real-world applications.

Our main contribution is introducing a system design and integration approach for a modular, high reconfigurable, mobile manipulation robot platform for manipulation in agriculture. 
The robot presented, named Hefty, is built upon Farm-ng's Amiga platform, which has been modified to incorporate a reconfigurable robot arm so that it can be used to perform various manipulation tasks. 
Hefty is specifically designed to maximize modularity and reconfigurability across all aspects of its design, including mobility, sensing, power, computing, and mounting features like sensors and manipulator.
Hefty's default configuration is specifically for research and development of crop-agnostic autonomous navigation, but it is easily modified for manipulation. 

The contributions of this paper are:
\begin{itemize}
    \item Introducing a system design and integration approach for a utility robot for agricultural manipulation that is modular and reconfigurable.
    \item Presenting multiple examples that demonstrate the usage of the robot is research efforts that range from navigation, near-ground manipulation, and over the canopy manipulation.
\end{itemize}

The rest of the manuscript is structured as follows: section \ref{sec:platform} discussed the design objects, and the realizations of said objectives through the design of mechanical, electrical components of the system. Decisions about computing, programming interfaces, and sensors are also presented in the same section. Section \ref{sec:demos} presents the summary of demonstration tasks, and the performance in those tasks. Section \ref{sec:conclusion} is the discussion section.

\begin{figure}[t]
  \centering
  \includegraphics[width=0.295\textwidth]{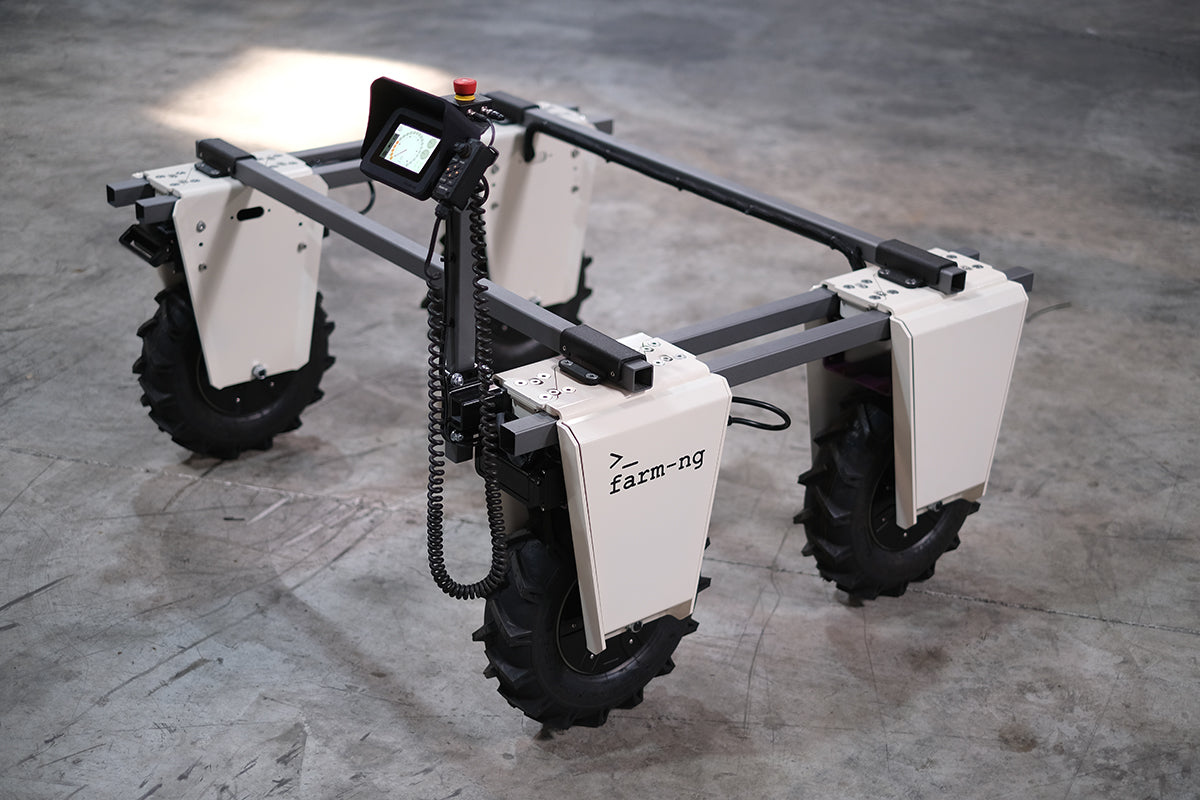}    
  \caption{Default Amiga robot configuration, as provided by Farm-ng, before all the modifications to accommodate robot manipulation configurations.}\label{fig:amiga-default}
\end{figure}

\section{Robot Platform}\label{sec:platform}
The design objectives for a reconfigurable modular robot for mobile manipulation in agriculture, later named Hefty, were established through the Artificial Intelligence for Resilient Agriculture (AIIRA) inter-disciplinary research initiative. AIIRA is a multi-institution research effort developing practical artificial intelligence tools for agriculture. One of AIIRA's primary aims is to generate morphologically and biochemically accurate virtual crops using real-world data. Digital twins, as they are called, could potentially be the driving model for data-driven decision-making in plant breeding and predictive plant health monitoring. This research initiative requires immense data collection, sensor deployment, and analysis that only humans typically do. In particular, the sensor deployments and data collection involve long hours of field visits and complicated sensor handling and reading.

Through discussions with collaborating agronomists, sensor developers, and roboticists, the design specifications of a utility research platform were defined as follows:
\begin{figure}[t]
\centering
\parbox{\figrasterwd}{
  \parbox{.175\figrasterwd}{
    \subcaptionbox{}{\includegraphics[width=\hsize]{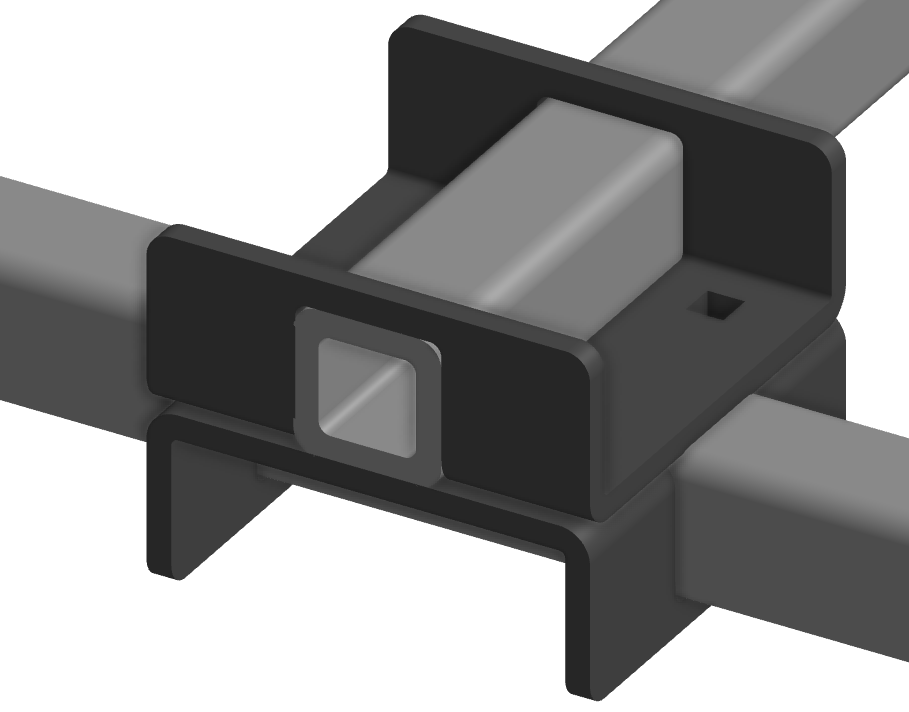}}
    \vskip1em
    \subcaptionbox{}{\includegraphics[width=\hsize]{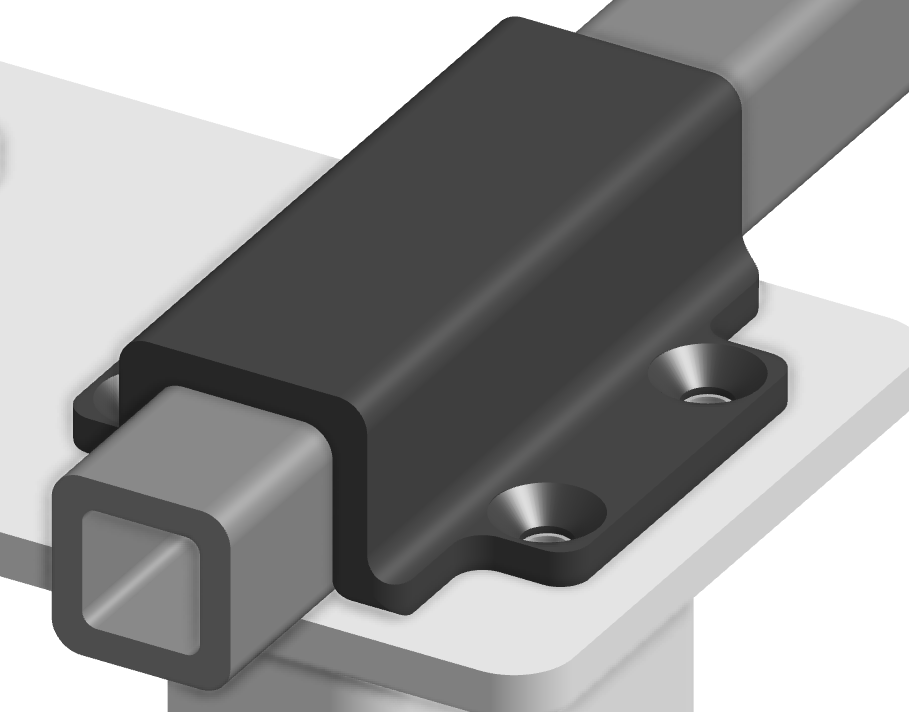}}\label{fig:cross-clamp}
  }
  \parbox{.15\figrasterwd}{
    \subcaptionbox{}{\includegraphics[width=\hsize, height=6cm]{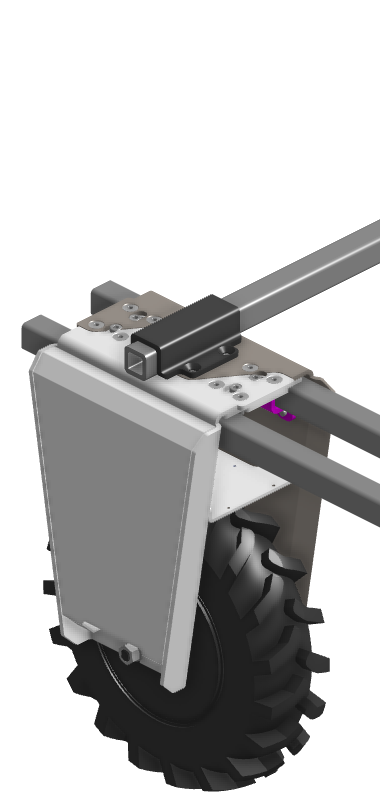}}
  }
  \parbox{.15\figrasterwd}{
    \subcaptionbox{}{\includegraphics[width=\hsize, height=6cm]{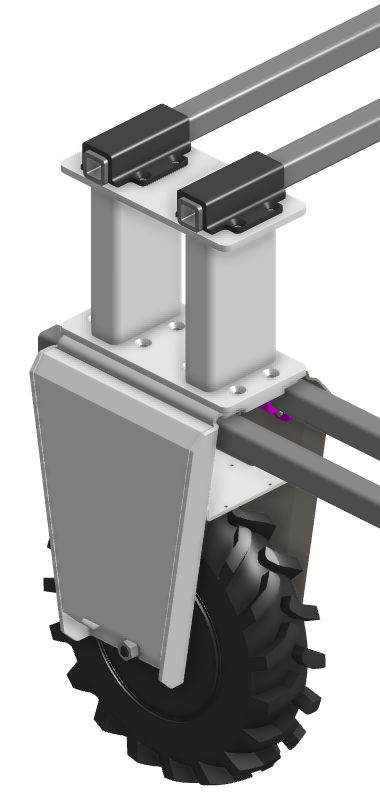}}
  }
}
\caption{The crossing beams and the single beams (shown in (a) and (b), respectively) make the Amiga's structural joinery robust and modular. Images (c) and (d) show how the same joinery is applied to mechanical risers to increase the height clearance.}\label{fig:hefty_features}
\end{figure}

\begin{itemize}
  \item Reconfigurability: the robot should be reconfigurable in (1) its mechanical form different crop sizes and field measurements and (2) fixture mounting to accommodate different sensor and manipulator configurations for researchers to determine what is best for their applications.
  \item Modularity: the robot should be built from modular units in five areas: mobility, sensing, power, computing, and fixture mounting. Modularity will allow all users to acquire and use modules that are relevant to their work. 
  \item Manipulation: the robot should be designed for manipulation tasks. Manipulation is one of the unique contributions of robotics compared to other types of mechanization; hence, the robot should be designed for research and use of manipulation in agriculture.
\end{itemize}
The target user for the robot platform includes researchers and practitioners, i.e., farmers and agronomists. Extending the target users to practitioners is essential in reducing the barriers to technology transfer because the researchers developing the technology and practitioners, like farmers, would share the same versatile, reconfigurable, and modular robot platform -- thus, the cutting-edge technology may be a configuration or module away from the end-user. 

\subsection{The Design}
The Hefty robot is built on a modified Amiga robot platform from Farm-ng\cite{Farmng}. The Amiga (shown in Fig.~\ref{fig:amiga-default}) is an established robust robot platform designed for plowing, weeding, and even manure spreading – it has a 500lb carrying capacity. The Amiga also comes with a custom programming interface.

Modifying the Amiga platform, instead of developing a completely custom solution, prioritizes solving critical design problems while taking advantage of existing solutions. In particular, the Amiga platform provides a lightweight, reconfigurable mobile platform that allows users to adjust the width to meet the required crop intra-line width. Its wheel modules are independently controllable via a CANBus interface, and the Power-over-Ethernet (PoE) cameras are accessible over a standard RJ45 water-proof interface. The Amiga's mechanical and electrical systems are also modular, and reconfigurable, and they are designed for the harsh operating conditions that typify open field agriculture. All these features make the Amiga a good base platform for developing robot for agricultural manipulation tasks.

\subsubsection{Structural Design}
The focus of the structural design was to establish a standard approach for increasing the robot clearance and mounting fixtures like manipulators and sensors. The Amiga solves this problem by using 1/4-inch square steel tubing and two types of clamping brackets, one for crossing bars and the other for single bars (shown in Fig~\ref{fig:hefty_features}(a) and Fig~\ref{fig:hefty_features}(b)). These modules are retained in the new design because they are robust, as evidenced by the Amiga's 500lb carrying capacity. Nonetheless, a new module is required to increase the height clearance from 23" to 32".
\begin{figure}[t]
  \centering
  \includegraphics[width=0.45\textwidth]{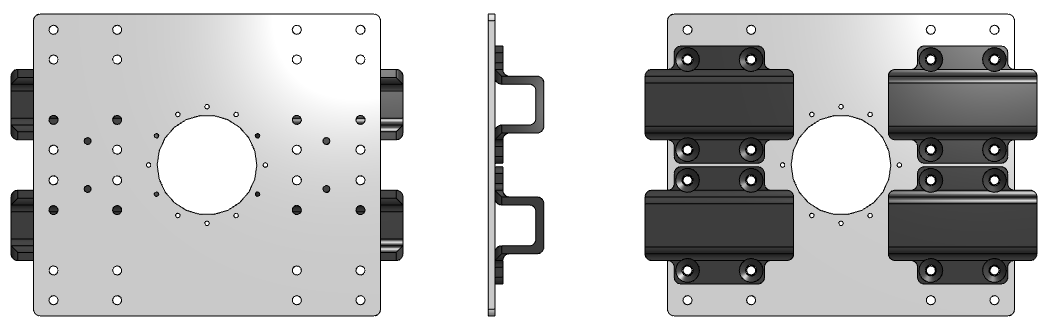}
  \caption{A closer look at the modular design of the manipulator mounting plate used in Fig.~\ref{fig:tasks} configuration examples. From the right, the images show the top, side and bottom orthographic views. The key modularity feature is using  the single-bar clamps to mount to any 1.5" parallel bars.}\label{fig:mounting-plate}
\end{figure}

To achieve the required height clearance of 32", the first structural update introduced fixed-height mechanical risers (shown in Fig.~\ref{fig:hefty_features}(d)). The key design features of the mechanical risers are the two mounting plates that use matching M8 bolt-pattern for mounting on the wheel modules, and the double cross bars (mounted using the single bar clamps). Two 4$\times$2-inch rectangular sections are welded between the two mounting plates. This design is robust enough to counter the stress and strains expected from increasing the height clearance of the robot.

The next challenge was designing a manipulator mounting module that allows reconfigurability -- particularly for reaching over and under the canopy. The solution was leveraging the parallel bars that feature throughout the structure with the mechanical risers. A manipulator mounting plate uses four clamps, as shown in Fig.~\ref{fig:mounting-plate}, was designed to mount anywhere along the parallel bars, and it only requires a matching bolt-pattern for the specific manipulator the user prefers. In the demostration tasks, the xArm6 6DOF arm from Ufactory was the preferred arm is it requires an M5 hexigonal threads pattern on a 110mm diameter. The mounting plate in Fig.~\ref{fig:mounting-plate} has a dodecagon thread pattern that increases the resolution of the robot base orientation options when mounting the manipulator.

\begin{figure}[t]
  \centering
  \includegraphics[width=0.1\textwidth]{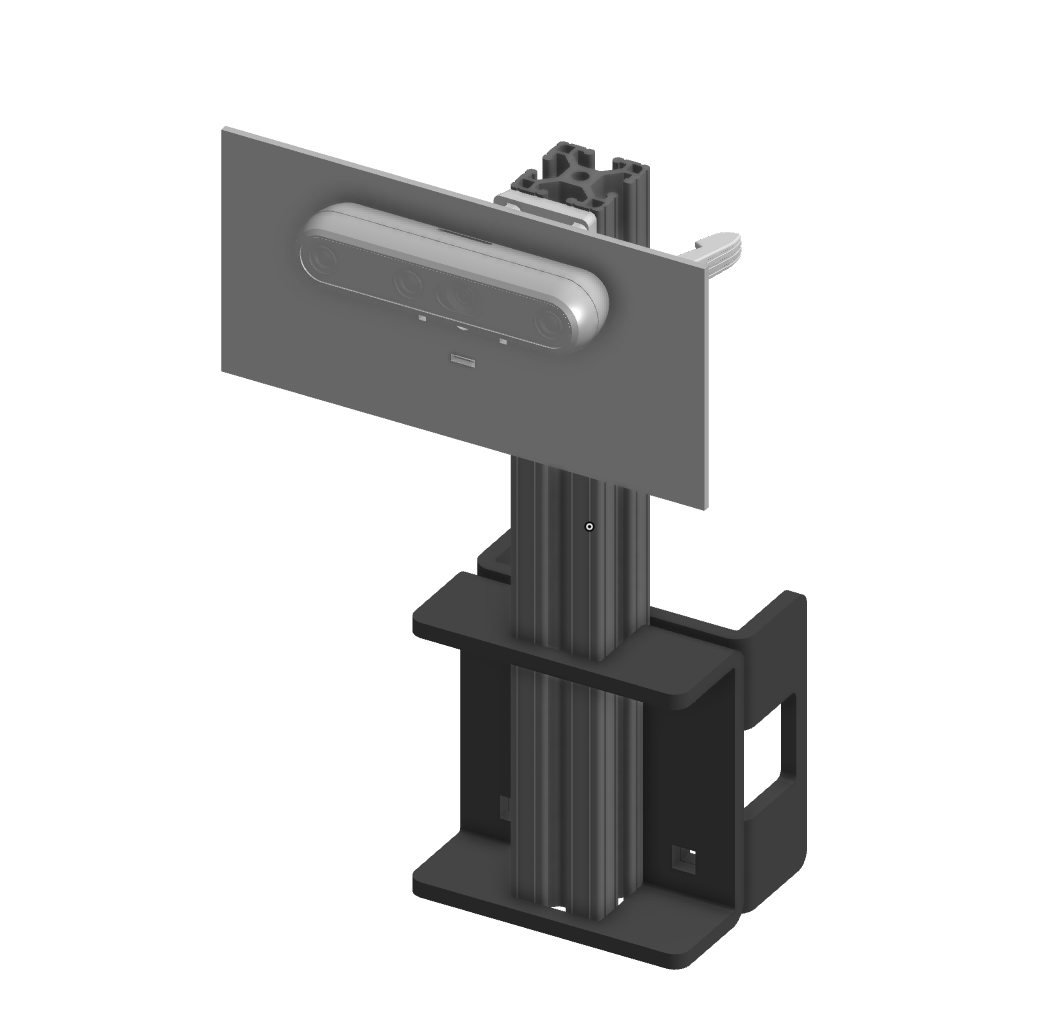}
  \includegraphics[width=0.1\textwidth]{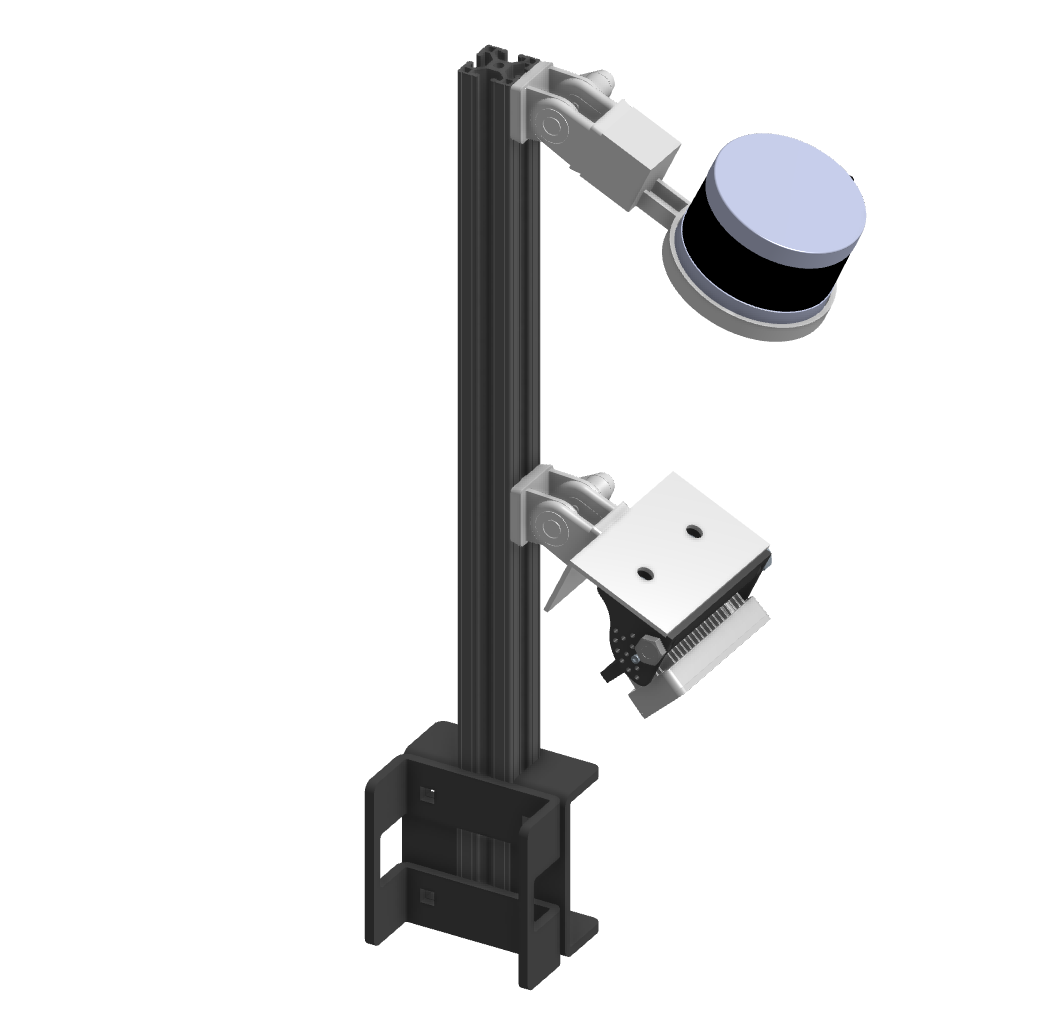}
  \includegraphics[width=0.1\textwidth]{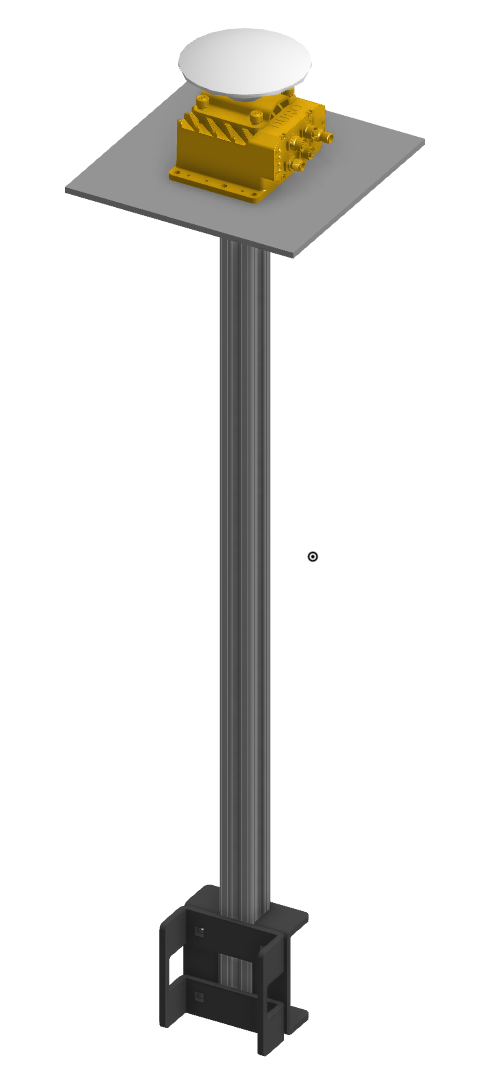}
  \includegraphics[width=0.1\textwidth]{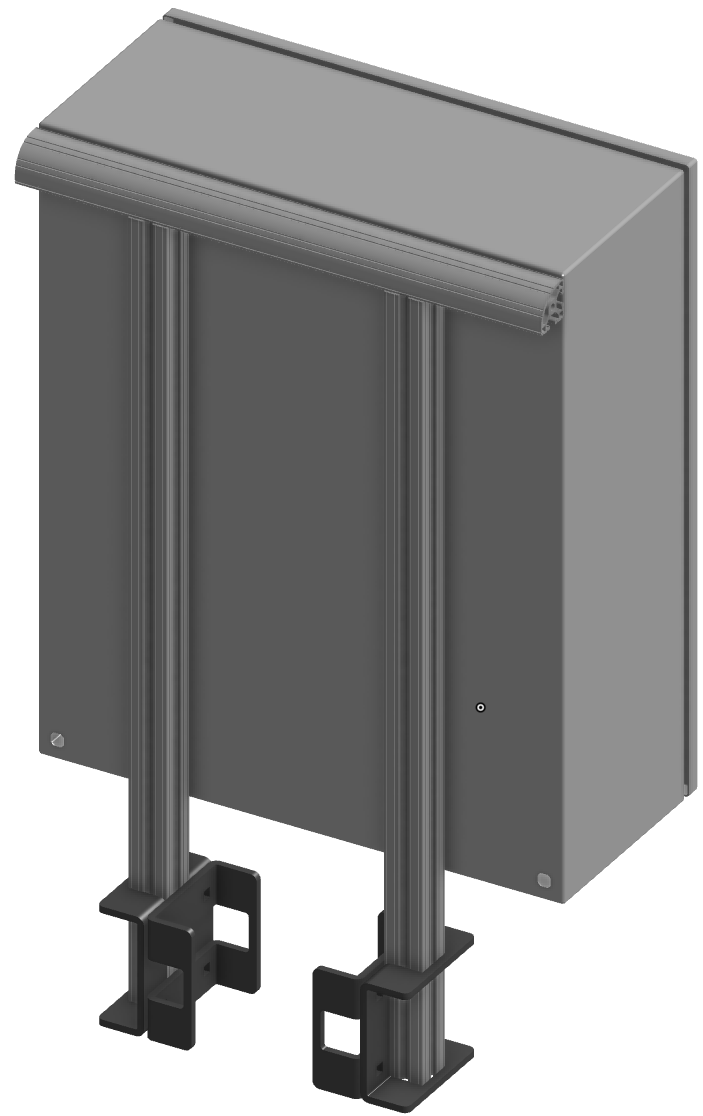}
  \caption{Examples of how  1.5in 80-20 Aluminum improves the modularity and reconfigurability of fixture mounting on the Hefty robot. From left: 3D models of a RealSense Camera (d455), an Oak-D W Pro camera below a Velodyne V16 Lidar, a SwiftNav GPS unit, and a weather-proof electric box housing the system computer and electrical system.}\label{fig:8020}
\end{figure}

Other fixtures, including the sensors and electrical box that houses the computer and power system, were mounted on the robot structure using 80-20 T-slotting extruded aluminum. Since the Amiga is built from 1.5" steel pipes, the 1.5" extruded aluminum is compatible with the existing clamping joinery, and it extends the modularity and reconfigurability of the robot, which is particularly essential for positioning sensors like lidars and cameras. Depending on the supplier, most extruded aluminum comes with additional fasteners and mechanical fittings -- the lidar and cameras in Fig.~\ref{fig:8020} use T-slotted channels for height adjustment and an angle adjustable fitting for changing the camera angle. The GPS unit and the weather-proof box were also mounted on the T-slotted aluminum primarily because aluminum is easier to machine than steel.

\subsubsection{Computing}\label{sec:compute}
The Hefty computer system is based on a consumer custom mini-ATX motherboard for a gaming computer. Using a custom-built computer allows the user to select their computing solution to fit their budget and computing needs. For Hefty, the mini-ATX motherboards prioritized maximizing the number of IO devices, primarily USB3 support, x86 processor, and NVIDIA GPU compatibility. 
Using an x86 processor ensured device driver support for the sensors and actuators, and the NVIDIA GPU guaranteed the deep learning models can be run onboard without any need for model compression. The deployed system was a ROG Z790 mini-ATX board (3$\times$USB-C, 6$\times$USB-A and 1$\times$Ethernet ports), with an Intel i9 139000 (24 Cores, 32 Threads) and an NVIDIA GeForce RTX 4070. While multi-processor CPUs are the standard, maximizing the number of cores was particularly essential for data logging operations which Hefty is expected to run most of the time.

\begin{figure}[ht]
  \centering              
  \includegraphics[width=0.5\textwidth]{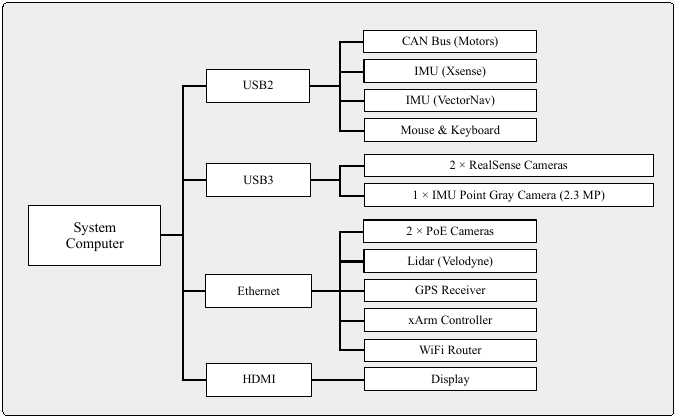}
  \caption{ This diagram outlines the communication protocols between the system computer and the sensors and actuators. Most devices use the USB protocol; matching the device with the correct USB version is essential for the best data throughput rates. }\label{fig:hefty_electrical}
\end{figure}

\subsubsection{Sensors \& Actuators}

To meet the diverse needs of its users, the robot was designed to handle a wide variety of sensor payloads. However, an autonomous navigation research sensor payload is the default sensor configuration, see Table~\ref{tab:default-payload} for a list of all the featured sensors. Besides the fact that autonomous navigation is a critical function for all autonomous task execution in agriculture, it also requires some of the most complex high-throughput sensor payloads that demonstrate the system's capability as a whole. Fig~\ref{fig:hefty_electrical} shows the computer-sensor-actuator communication network and the robot power supply.

Most of the sensors use USB communication and Ethernet. Of particular note, RealSense Cameras required direct-to-motherboard connections, instead of via a USB hub, for reliable performance. 

\begin{table}
\renewcommand{\arraystretch}{1.3}
\caption{List of default sensor and actuator payload for navigation research}\label{tab:default-payload}
\centering
\begin{tabular}{c||c|c}
\hline
  \bfseries Component & \bfseries \# & Comm. \\
\hline\hline
  RealSense d455 & 2 & USB-C \\
  Grasshopper 3.2MP & 1 & USB-A \\
  USB-CAN adapter & 1 & USB-A \\
  Xsense IMU & 1 & USB-A \\
  VectorNav IMU & 1 & USB-A \\
  Velodyne V16 Lidar & 1 & Ethernet \\
  PoE Cameras & 2 & Ethernet \\
  xArm Controller & 1 & Ethernet \\
  SwiftNav GPS & 1 & Ethernet \\
\hline
\end{tabular}
\end{table}

\subsubsection{Power System}
\begin{figure}[H]
  \centering              
  \includegraphics[width=0.5\textwidth]{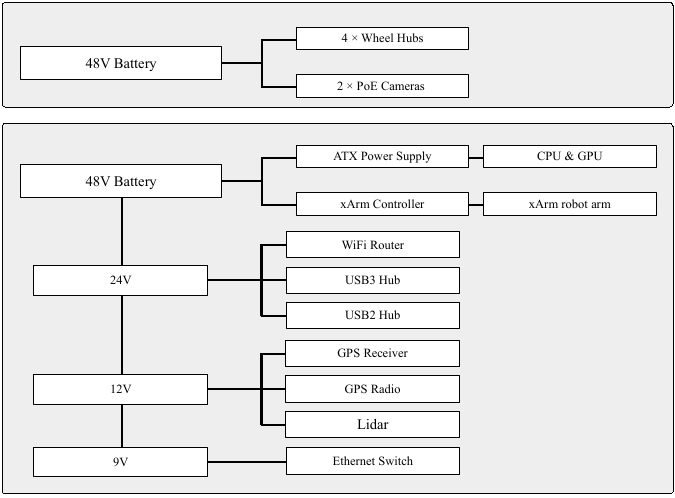}
  \caption{This illustrative diagram shows how the two power systems powering the robot are connected. The original power system (top) is retained and used to power the Amiga mobile base. The new system (bottom) powers the computer, sensors, and the manipulator. Any direct connections between two voltage levels represent a voltage down-regulation.}\label{fig:hefty_power}
\end{figure}

The presented robot uses two separate power supply circuits. Both power supplies use the Amiga's 48V 1.32kWh Lithium-Ion batteries with an in-built 30A fuse. As illustrated in Fig.~\ref{fig:hefty_power}, the first circuit powers the wheel hubs and the PoE cameras; this is the default Amiga power system that is not modified, except for removing devices like the Amiga computer. The batteries housed in the wheel hubs are connected in parallel; thus, one or more batteries can be connected simultaneously and are hot-swappable. The voltage regulators are in the wheel hubs or adjacent to the PoE injector switch.

The new circuit powers all the new devices like the computer system, sensors, actuators, and all the secondary devices like Ethernet switches, USB hubs, and a WiFi router. This circuit is powered using a single battery mounted directly below the electric box. Fig.~\ref{fig:hefty_power} shows all the provided voltages distributed using a DIN rail-based voltage regulator system occupying the bottom half of the electric box.

The computer and the arm controller came with voltage regulators, either in the form of an independent DC ATX power supply or an internal voltage regulation of the xArm controller; hence, they were directly connected to the batteries. Devices like the USB-CAN adapter and USB cameras were directly powered via their USB connections.

\subsubsection{Programming Interface}
The Hefty robot uses ROS as the programming interface. ROS is a widely used open-source standard middleware for modularizing the software implementation of data processing and control algorithms, and it also simplifies operations like sensor data logging, system monitoring, and communication with sensors and actuators. Additionally, many sensors and actuators come with ROS drivers from their suppliers or open-source developed. 

\section{Demonstration Tasks}\label{sec:demos}



\subsection{Navigation}

The navigation research team is working on two problems: barn-to-field and in-row navigation. Barn-to-field navigation involves moving the robot to and fro its storage place, likely a barn, and the deployment field, where it executes crop-specific tasks. In-row navigation, on the other hand, is about navigating the robot within the plants, typically supporting an agricultural task like insect scouting, pruning, or harvesting. Both tasks depend on real-time kinematics RTK-GPS, lidar, and vision sensors to solve the relevant perception, planning, and control challenges. The computing capability of the robot and its reconfigurability give the researchers the flexibility to experiment with different algorithms and sensor configurations.

In our current implementation, the barn-to-field navigation works using a two-step process. First, photogrammetry software is used to construct a centimeter-level accurate field map using images collected using a drone -- the drone images contain ground control points with known RTK-GPS coordinates. The user specifies the robot's path by drawing on the field map, and the robot follows this path using a model predictive controller (MPC) trajectory tracking controller. No significant hardware modifications are required except for removing irrelevant sensors and fixtures and ensuring that the perception sensors are not obstructed.

The goal of in-row navigation is to create crop-agnostic algorithms that move the robot safely with minimal user tuning. Besides new algorithms, this problem requires the perception sensors (cameras and lidar) to be positioned for maximum coverage and details. Hence, the primary modifications consist of placing the lidar and camera the furthest in front of the robot and as high as possible without generating significant flexural oscillation in the mounting beam that may cause the images to be blurry. Most of the work is still in development. However, the current implementation of vision-based navigation is performing well in sparse weed-free soybeans, and lidar-based navigation is still under development.

\subsection{Insect Scouting}

Insect scouting, in general, is not only about moving around inspecting for the presence of agricultural pests but also includes monitoring the health of beneficial insects like pollinating bees. It is time-consuming and tiresome. Robot-based scouting provides numerous advantages, including consistent performance, data logging, and potential real-time analysis. In this research, the insect scouting team used an in-hand active lighting camera \cite{silwalRobustIlluminationInvariantCamera2021} to take consistent, well-lit images and used InsectNect\cite{chiranjeeviDeepLearningPowered2023} to detect and classify the present insects. InsectNet is a RegNet-based deep learning model, also running on the RTX4070 GPU, that classifies up to 2500 insect species.
\begin{figure}[h]
    \centering
    \includegraphics[width=0.495\textwidth]{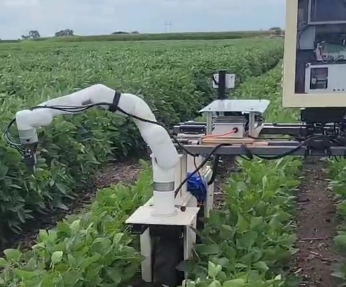}
    \caption{The insect scouting task required over-the-canopy imaging with a custom active lighting in-hand stereo camera. The white structure next to the manipulator houses the high-voltage power supply for the camera flash system.}\label{fig:task-over-canopy}
\end{figure}
The main modification for this task was mounting the robot arm for maximum area coverage when imaging the soybean plant foliage. Other factors for better image quality were the camera's focal length and avoiding over-saturation when the camera is too close to the target. After all considerations, the best mounting position was determined to be right-side up on a mounting plate installed on two bars protruding backward on the back-left wheel hub, as shown in \textcolor{black}{Fig.~\ref{fig:task-over-canopy}}.

The camera system requires 3 USB connections and a high-voltage power supply to power the flash. Hence, a platform was added to the robot structure to hold an additional Amiga battery, the flash power supply circuitry, and an additional USB hub with an actively powered USB cable to ensure the resilience of the camera-to-computer high throughput when transmitting image data. The new platform had the advantage of effectively modularizing the stereo-camera system for easier installation and removal.

\subsection{Sensor Insertion}

This research effort is an example of using robot manipulation to deploy sensors that humans would otherwise deploy. In this case, a small nitrogen sensor \cite{ibrahimPlantaNitrateSensor2022,jiaoInPlantaNitrateDetection2019}, with a shape reminiscent of a 1" square-cut nail with a sharp edge, is installed by stabbing it into the cornstalk so that its transducers detect the nitrogen level in the plant's sap. The sensor insertion team developed a custom end-effector with a Realsense camera for identifying the target cornstalk, a linear actuator with a sensor holding mechanism for pushing into the cornstalk, and a microcontroller for controlling the linear actuator.
\begin{figure}
    \centering
    \includegraphics[width=0.45\textwidth]{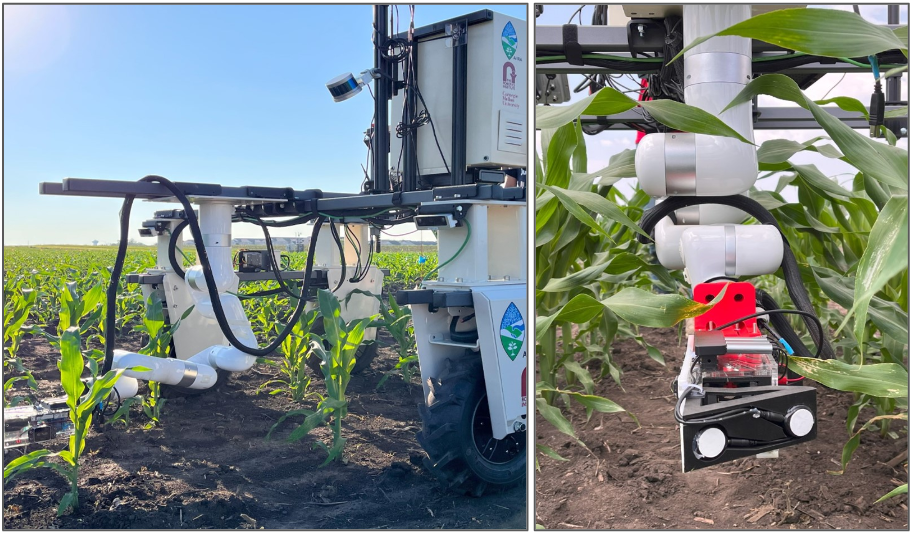}
    \caption{For cornstalk sensor insertion, the manipulator is mounted on bars protruding ahead of the robot, and upside down to reach as close to the ground as possible.}\label{fig:task-corn}
\end{figure}

The sensor is inserted into the stalk only a few inches above the ground. Hence, the first modification is mounting the arm in an upside-down configuration so its most manipulable workspace is as close to the ground as possible. As shown in  \textcolor{black}{Fig.~\ref{fig:task-corn}}, the arm was mounted on two bars extending straight ahead, front-center of the robot. The corn was planted 30" between lines, and the robot was 60" wide; hence, the arm hangs between the two corn lines the robot straddles over. The second modification was about electrical wiring: two new USB connections were made to the RealSense camera and the microcontroller, and an additional 12V line was extended from the DIN-rail power distribution panel to the linear actuator.

The cornstalk detection algorithm was based on a Mask-R-CNN deep learning model for image segmentation. Specifically, the model segments the cornstalks using the RGB data from the RGBD from the Realsense camera. The corresponding depth information is used to locate the cornstalks so the arm can reach and insert the sensor. The deep learning model ran on the GPU unit.

\subsection{Pepper Harvesting}

\begin{figure}
    \centering
    \includegraphics[width=0.25\textwidth]{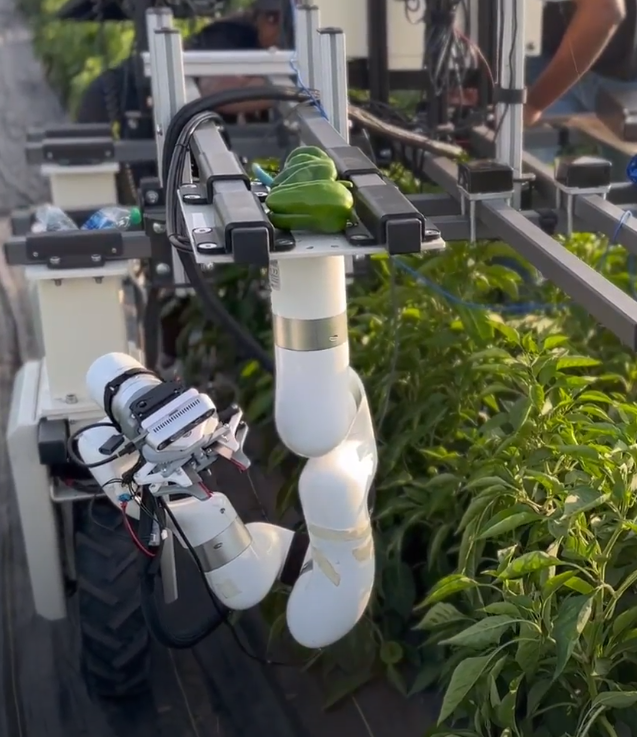}
    \caption{This is the manipulator mounting configuration for pepper harvesting. It is off-center, and elevated using an 80-20 aluminum structure to avoid colliding with the plants, and maximize reachability into the side of the plant canopy.}\label{fig:task-pepper}
\end{figure}

Pepper harvesting is also another example of a task left to humans due to its demand for dexterity and safe handling of both the produce and the plant. In this project, the robot was used to evaluate the effectiveness of a semi-autonomous harvesting system in a real-world deployment test. Broadly, the project aims to develop an autonomous pepper-harvesting robot to address labor shortages in the agricultural sector. To this end, the end-effector, shown in Fig.~\ref{fig:task-pepper}, was developed for grasping and cutting the peduncle supporting the pepper. It features two Dynamixel motors and a Realsense camera connected to the computer by USB, and the motors are powered through a 12V regulator in the computer power distribution system.

Pepper harvesting also requires an upside-down manipulator. Compared to the sensor insertion task, the mounting plate is placed higher to avoid colliding with the plants and to raise its dextrous workspace to match the likely height distribution of peppers in the plant canopy. A 1.5" square 80-20 extruded aluminum platform was constructed and attached to the robot structure using the Amiga clamping brackets (see Fig.~\ref{fig:task-pepper})) to provide a robust support frame for the manipulator mounting plates.

This pepper harvesting system is semi-autonomous because the gripper and arm were partly controlled by teleoperation. However, a YOLO-based deep learning model refined for the recognition of pepper fruits was also used in an autonomous harvesting algorithm that uses visual servoing.

\subsection{Fighting Lantern Flies}
\begin{figure}[h]
    \centering
    \includegraphics[width=0.25\textwidth]{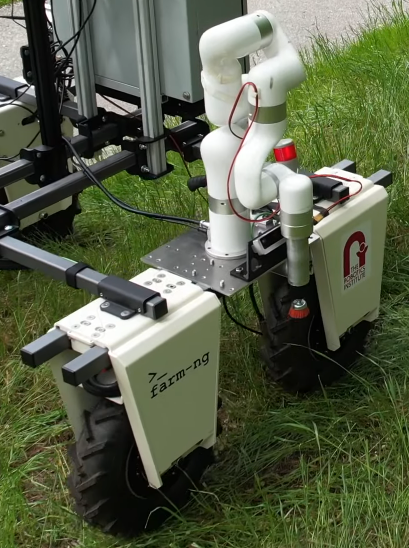}
    \caption{The configuration in the image was chosen to reach close to the ground and higher up on tree trunks where egg masses are often located. The end-effector is a modified hand drill with a wire brush attachment.}\label{fig:task-lantern-fly}
\end{figure}

In the configuration shown in Fig.~\ref{fig:task-lantern-fly}, Hefty was used to destroy lantern fly egg masses. Lantern flies are moth-like insects that are inversive in the northeast of the United States and spreading. They cause significant damage to vineyards and orchards, and they are generally an annoyance in most places they affect. To fight the spread of lantern flies, the prototype presented in Fig.~\ref{fig:task-lantern-fly} was developed to use image recognition to identify lantern fly egg masses and to use a motorized abrasive brush end-effector to exterminate them.

The modifications for this task were relatively minimal compared to most tasks. The risers were not used because the target workspace was about 1ft off the ground and higher. The manipulator was between the front and back wheels to the left side. A PoE camera was mounted at the base of the manipulator to capture the robot target workspace, and it also ran, on-device, a YOLO object recognition model refined on open-source lantern fly egg mass images sourced from iNaturalist\cite{INaturalist}.

\section{Conclusions and Future Work}\label{sec:conclusion}
This paper introduces a utility modular and reconfigurable mobile manipulation robot for agricultural applications. This robot, named Hefty, is built on Farm-ng's Amiga mobile robot, and it is designed to be modular in five aspects: mobility, sensing, power, computing, and mounting fixtures. Its utility is demonstrated in four research efforts that include navigation, cornstalk sensor insertion, insect scouting, and pepper harvesting.
Through the presented tests, a few limitations were presented. The first is the battery capacity of the computer system -- unlike the mobile platform that can take more than one battery, the current system only accepts one; hence, future modifications will include adding more battery capacity. Another limitation of the system is that all teams were required to share the same operating system installation. This caused clashes in preferred device driver versions and software dependencies. Hence, more effort will be invested in establishing a workflow that leverages system virtualization and containerization to isolate development and runtime environments across research teams. 
Future upgrades include multi-arm support for complex manipulation tasks like interactive inspection. A new research effort is currently exploring AI-driven design for determining the optimal configuration for the mobile platform and the manipulator, given a task description, including constructing modular manipulators from submodules.

\section*{Acknowledgment}
This work was supported in part by NSF Robust Intelligence 1956163, and NSF/USDA-NIFA AIIRA AI Research Institute 2021-67021-35329.

\bibliographystyle{ieeetr}
\bibliography{ag_rsch_robots}

\begin{thebibliography}{10}

\bibitem{oliveiraAdvancesAgricultureRobotics2021}
L.~F.~P. Oliveira, A.~P. Moreira, and M.~F. Silva, ``Advances in {{Agriculture
  Robotics}}: {{A State-of-the-Art Review}} and {{Challenges Ahead}},'' {\em
  Robotics}, vol.~10, p.~52, Mar. 2021.

\bibitem{srinivasanDesignAutonomousSeed2016}
N.~Srinivasan, P.~Prabhu, S.~S. Smruthi, N.~V. Sivaraman, S.~J. Gladwin,
  R.~Rajavel, and A.~R. Natarajan, ``Design of an autonomous seed planting
  robot,'' in {\em 2016 {{IEEE Region}} 10 {{Humanitarian Technology
  Conference}} ({{R10-HTC}})}, pp.~1--4, Dec. 2016.

\bibitem{haiboStudyExperimentWheat2015}
L.~Haibo, D.~Shuliang, L.~Zunmin, and Y.~Chuijie, ``Study and {{Experiment}} on
  a {{Wheat Precision Seeding Robot}},'' {\em Journal of Robotics}, vol.~2015,
  pp.~1--9, 2015.

\bibitem{benderHighresolutionMultimodalData2020}
A.~Bender, B.~Whelan, and S.~Sukkarieh, ``A high-resolution, multimodal data
  set for agricultural robotics: {{A Ladybird}}'s-eye view of {{Brassica}},''
  {\em Journal of Field Robotics}, vol.~37, no.~1, pp.~73--96, 2020.

\bibitem{EcorobotixSmartSpraying}
``Ecorobotix : {{Smart}} spraying for ultra-localised treatments..''
  https://ecorobotix.com/en/.

\bibitem{ovchinnikovKinematicStudyWeeding2019}
A.~S. Ovchinnikov, V.~S. Bocharnikov, N.~S. Vorob'yeva, and A.~G. Ivanov,
  ``Kinematic study of the weeding robot,'' {\em IOP Conference Series:
  Materials Science and Engineering}, vol.~489, p.~012056, Mar. 2019.

\bibitem{ruckelshausenBoniRobAutonomousFielda}
A.~Ruckelshausen, P.~Biber, M.~Dorna, H.~Gremmes, R.~Klose, A.~Linz, R.~Rahe,
  R.~Resch, M.~Thiel, D.~Trautz, and U.~Weiss, ``{{BoniRob--an}} autonomous
  field robot platform for individual plant phenotyping,'' {\em Precision
  agriculture}, vol.~9, no.~841, p.~1, 2009.

\bibitem{CarbonRobotics}
``Carbon {{Robotics}}.'' https://carbonrobotics.com.

\bibitem{vishnurajendranSelectiveHarvestingRobots}
S.~Vishnu~Rajendran, U.~K. Lincoln, B.~Debnath, S.~Mghames, W.~Mandil,
  S.~Parsa, S.~Parsons, and A.~Ghalamzan, ``Selective {{Harvesting Robots}}:
  {{A Review}},''

\bibitem{silwalDesignIntegrationField2017}
A.~Silwal, J.~R. Davidson, M.~Karkee, C.~Mo, Q.~Zhang, and K.~Lewis, ``Design,
  integration, and field evaluation of a robotic apple harvester,'' {\em
  Journal of Field Robotics}, vol.~34, no.~6, pp.~1140--1159, 2017.

\bibitem{davidsonProofofconceptRoboticApple2016}
J.~R. Davidson, A.~Silwal, C.~J. Hohimer, M.~Karkee, C.~Mo, and Q.~Zhang,
  ``Proof-of-concept of a robotic apple harvester,'' in {\em 2016
  {{IEEE}}/{{RSJ International Conference}} on {{Intelligent Robots}} and
  {{Systems}} ({{IROS}})}, pp.~634--639, Oct. 2016.

\bibitem{hayashiRoboticHarvestingSystem2002}
S.~Hayashi, K.~Ganno, Y.~Ishii, and I.~Tanaka, ``Robotic harvesting system for
  eggplants,'' {\em Japan Agricultural Research Quarterly: JARQ}, vol.~36,
  no.~3, pp.~163--168, 2002.

\bibitem{sepulvedaRoboticAubergineHarvesting2020}
D.~Sepulveda, R.~Fernandez, E.~Navas, M.~Armada, and P.~{Gonzalez-De-Santos},
  ``Robotic {{Aubergine Harvesting Using Dual-Arm Manipulation}},'' {\em IEEE
  Access}, vol.~8, pp.~121889--121904, 2020.

\bibitem{yuRealTimeVisualLocalization2020}
Y.~Yu, K.~Zhang, H.~Liu, L.~Yang, and D.~Zhang, ``Real-{{Time Visual
  Localization}} of the {{Picking Points}} for a {{Ridge-Planting Strawberry
  Harvesting Robot}},'' {\em IEEE Access}, vol.~8, pp.~116556--116568, 2020.

\bibitem{bacHarvestingRobotsHighvalue2014}
C.~W. Bac, E.~J. {van Henten}, J.~Hemming, and Y.~Edan, ``Harvesting {{Robots}}
  for {{High-value Crops}}: {{State-of-the-art Review}} and {{Challenges
  Ahead}},'' {\em Journal of Field Robotics}, vol.~31, no.~6, pp.~888--911,
  2014.

\bibitem{schutzModularRobotSystema}
C.~Sch{\"u}tz, J.~Pfaff, J.~Baur, T.~Buschmann, and H.~Ulbrich, ``A {{Modular
  Robot System}} for {{Agricultural Applications}},''

\bibitem{levinDesignTaskBasedModular2016}
M.~Levin and A.~Degani, ``Design of a {{Task-Based Modular Re-Configurable
  Agricultural Robot}},'' {\em IFAC-PapersOnLine}, vol.~49, pp.~184--189, Jan.
  2016.

\bibitem{levinConceptualFrameworkOptimization2019}
M.~Levin and A.~Degani, ``A conceptual framework and optimization for a
  task-based modular harvesting manipulator,'' {\em Computers and Electronics
  in Agriculture}, vol.~166, p.~104987, Nov. 2019.

\bibitem{atefiRoboticTechnologiesHighThroughput2021}
A.~Atefi, Y.~Ge, S.~Pitla, and J.~Schnable, ``Robotic {{Technologies}} for
  {{High-Throughput Plant Phenotyping}}: {{Contemporary Reviews}} and {{Future
  Perspectives}},'' {\em Frontiers in Plant Science}, vol.~12, 2021.

\bibitem{zhangFieldPhenotypingRobot2016}
J.~Zhang, L.~Gong, C.~Liu, Y.~Huang, D.~Zhang, and Z.~Yuan, ``Field
  {{Phenotyping Robot Design}} and {{Validation}} for the {{Crop Breeding}},''
  {\em IFAC-PapersOnLine}, vol.~49, pp.~281--286, Jan. 2016.

\bibitem{xiangFieldBasedRobotic2023}
L.~Xiang, J.~Gai, Y.~Bao, J.~Yu, P.~S. Schnable, and L.~Tang, ``Field-based
  robotic leaf angle detection and characterization of maize plants using
  stereo vision and deep convolutional neural networks,'' {\em Journal of Field
  Robotics}, vol.~40, pp.~1034--1053, Aug. 2023.

\bibitem{baoFieldbasedArchitecturalTraits2019}
Y.~Bao, L.~Tang, S.~Srinivasan, and P.~S. Schnable, ``Field-based architectural
  traits characterisation of maize plant using time-of-flight {{3D}} imaging,''
  {\em Biosystems Engineering}, vol.~178, pp.~86--101, 2019.

\bibitem{alenyaRoboticLeafProbing2012}
G.~Aleny{\`a}, B.~Dellen, S.~Foix, and C.~Torras, ``Robotic leaf probing via
  segmentation of range data into surface patches,'' 2012.

\bibitem{atefiVivoHumanlikeRobotic2019}
A.~Atefi, Y.~Ge, S.~Pitla, and J.~Schnable, ``In vivo human-like robotic
  phenotyping of leaf traits in maize and sorghum in greenhouse,'' {\em
  Computers and Electronics in Agriculture}, vol.~163, p.~104854, Aug. 2019.

\bibitem{mueller-simRobotanistGroundbasedAgricultural2017}
T.~{Mueller-Sim}, M.~Jenkins, J.~Abel, and G.~Kantor, ``The {{Robotanist}}:
  {{A}} ground-based agricultural robot for high-throughput crop phenotyping,''
  in {\em 2017 {{IEEE International Conference}} on {{Robotics}} and
  {{Automation}} ({{ICRA}})}, ({Singapore, Singapore}), pp.~3634--3639, {IEEE},
  May 2017.

\bibitem{abelInfieldroboticLeafGrasping2018}
J.~Abel, ``In-fieldrobotic leaf grasping and automated crop spectroscopy,''
  {\em Carnegie Mellon University: Pittsburgh}, 2018.

\bibitem{colucciKinematicModelingMotion2022}
G.~Colucci, A.~Botta, L.~Tagliavini, P.~Cavallone, L.~Baglieri, and G.~Quaglia,
  ``Kinematic {{Modeling}} and {{Motion Planning}} of the {{Mobile Manipulator
  Agri}}.{{Q}} for {{Precision Agriculture}},'' {\em Machines}, vol.~10,
  p.~321, May 2022.

\bibitem{silwalBumblebeePathFully2022}
A.~Silwal, F.~Yandun, A.~Nellithimaru, T.~Bates, and G.~Kantor, ``Bumblebee:
  {{A Path Towards Fully Autonomous Robotic Vine Pruning}},'' {\em Field
  Robotics}, vol.~2, pp.~1661--1696, Mar. 2022.

\bibitem{grimstadThorvaldIIAgricultural2017}
L.~Grimstad and P.~J. From, ``The {{Thorvald II Agricultural Robotic
  System}},'' {\em Robotics}, vol.~6, p.~24, Dec. 2017.

\bibitem{Farmng}
``Farm-ng.'' https://farm-ng.squarespace.com.

\bibitem{silwalRobustIlluminationInvariantCamera2021}
A.~Silwal, T.~Parhar, F.~Yandun, H.~Baweja, and G.~Kantor, ``A {{Robust
  Illumination-Invariant Camera System}} for {{Agricultural Applications}},''
  in {\em 2021 {{IEEE}}/{{RSJ International Conference}} on {{Intelligent
  Robots}} and {{Systems}} ({{IROS}})}, pp.~3292--3298, Sept. 2021.

\bibitem{chiranjeeviDeepLearningPowered2023}
S.~Chiranjeevi, M.~Sadaati, Z.~K. Deng, J.~Koushik, T.~Z. Jubery, D.~Mueller,
  M.~E.~O. Neal, N.~Merchant, A.~Singh, A.~K. Singh, S.~Sarkar, A.~Singh, and
  B.~Ganapathysubramanian, ``Deep learning powered real-time identification of
  insects using citizen science data,'' June 2023.

\bibitem{ibrahimPlantaNitrateSensor2022}
H.~Ibrahim, S.~Yin, S.~Moru, Y.~Zhu, M.~J. Castellano, and L.~Dong, ``In
  {{Planta Nitrate Sensor Using}} a {{Photosensitive Epoxy Bioresin}},'' {\em
  ACS Applied Materials \& Interfaces}, vol.~14, pp.~25949--25961, June 2022.

\bibitem{jiaoInPlantaNitrateDetection2019}
Y.~Jiao, X.~Wang, Y.~Chen, M.~J. Castellano, J.~C. Schnable, P.~S. Schnable,
  and L.~Dong, ``In-{{Planta Nitrate Detection Using Insertable Plant
  Microsensor}},'' in {\em 2019 20th {{International Conference}} on
  {{Solid-State Sensors}}, {{Actuators}} and {{Microsystems}} \& {{Eurosensors
  XXXIII}} ({{TRANSDUCERS}} \& {{EUROSENSORS XXXIII}})}, pp.~37--40, June 2019.

\bibitem{INaturalist}
``{{iNaturalist}}.'' https://www.inaturalist.org/.

\end{thebibliography}
\end{document}